\documentclass[sigconf]{acmart}
\AtBeginDocument{%
  }

\setcopyright{acmlicensed}
\copyrightyear{2025} 
\acmYear{2025} 
\setcopyright{acmlicensed}\acmConference[MMAsia '25]{ACM Multimedia
 Asia}{December 9--12, 2025}{Kuala Lumpur, Malaysia}
 \acmBooktitle{ACM Multimedia Asia (MMAsia '25), December 9--12, 2025,
 Kuala Lumpur, Malaysia}
 \acmDOI{10.1145/3743093.3770955}
 \acmISBN{979-8-4007-2005-5/2025/12}

\begin{document}

\title{GAOT: Generating Articulated Objects Through Text-Guided Diffusion Models}

\author{Hao Sun}
\authornote{Both authors contributed equally to this research.}
\email{sunhaohit@stu.hit.edu.cn}
\affiliation{%
  \institution{Harbin Institute of Technology}
  \city{Harbin}
  \state{Heilongjiang}
  \country{China}
}
\orcid{0009-0003-2119-7199}

\author{Lei Fan}
\authornotemark[1]
\email{lei.fan1@unsw.edu.au}
\affiliation{%
  \institution{University of New South Wales}
  \city{Sydney}
  \state{NSW}
  \country{Australia}
}
\orcid{0000-0001-9472-7152}

\author{Donglin Di}
\email{donglin.ddl@gmail.com}
\affiliation{%
  \institution{Li Auto}
  \city{Beijing}
  \country{China}
}
\orcid{0000-0002-2270-3378}

\author{Shaohui Liu}
\email{shliu@hit.edu.cn}
\affiliation{%
  \institution{Harbin Institute of Technology}
  \city{Harbin}
  \state{Heilongjiang}
  \country{China}
}
\orcid{0000-0002-1810-5412}

\renewcommand{\shortauthors}{Sun et al.}

\begin{abstract}
Articulated object generation has seen increasing advancements, yet existing models often lack the ability to be conditioned on text prompts. To address the significant gap between textual descriptions and 3D articulated object representations, we propose GAOT, a three-phase framework that generates articulated objects from text prompts, leveraging diffusion models and hypergraph learning in a three-step process.  

First, we fine-tune a point cloud generation model to produce a coarse representation of objects from text prompts. Given the inherent connection between articulated objects and graph structures, we design a hypergraph-based learning method to refine these coarse representations, representing object parts as graph vertices. Finally, leveraging a diffusion model, the joints of articulated objects—represented as graph edges—are generated based on the object parts.
Extensive qualitative and quantitative experiments on the PartNet-Mobility dataset demonstrate the effectiveness of our approach, achieving superior performance over previous methods. 
\end{abstract}



\maketitle

\section{Introduction}
\label{sec:intro}

Articulated objects are characterized as composite entities consisting of multiple rigid segments interconnected by joints that facilitate rotational or translational movements. 
The primary objective of the generation of articulated objects is to comprehend the shape and dynamics of the articulated components, to represent the geometry and articulatory capabilities of the object's constituent parts, and to construct realistic models that accurately depict articulated objects as they exist in the physical world. Articulated object generation has broad applications in robotics~\cite{flowbot3d,universalmanipulation} and the research field of digital twins~\cite{dt3,Acdc}. 
Traditional methods of obtaining articulated objects include hand-made (via CAD) and realistic object scanning~\cite{Akb-48}, which are often time-consuming and costly.

Early studies \cite{pekelny2008} reconstructed articulated objects from depth video or a single snapshot \cite{mu2021sdf} using point cloud representations or Signed Distance Functions (SDF) representations. However, these methods primarily modeled articulated objects but could not generate articulated objects. With the growing application of deep learning~\cite{fan2025grainbrain,tang2025prototype,fan2024patch} and generative models~\cite{fan2022fast,fan2025manta,yin2025grpose}, especially diffusion models \cite{DDPM,ma2025adams,feng2025ditalker}, there is a strong drive to develop automated computational methods for generating articulated objects. Some research efforts have leveraged these 3D representation methodologies to create articulated objects, including notable works such as NAP~\cite{NAP} and cage~\cite{cage}.

\begin{figure}[t]
  \centering
   \includegraphics[width=\linewidth]{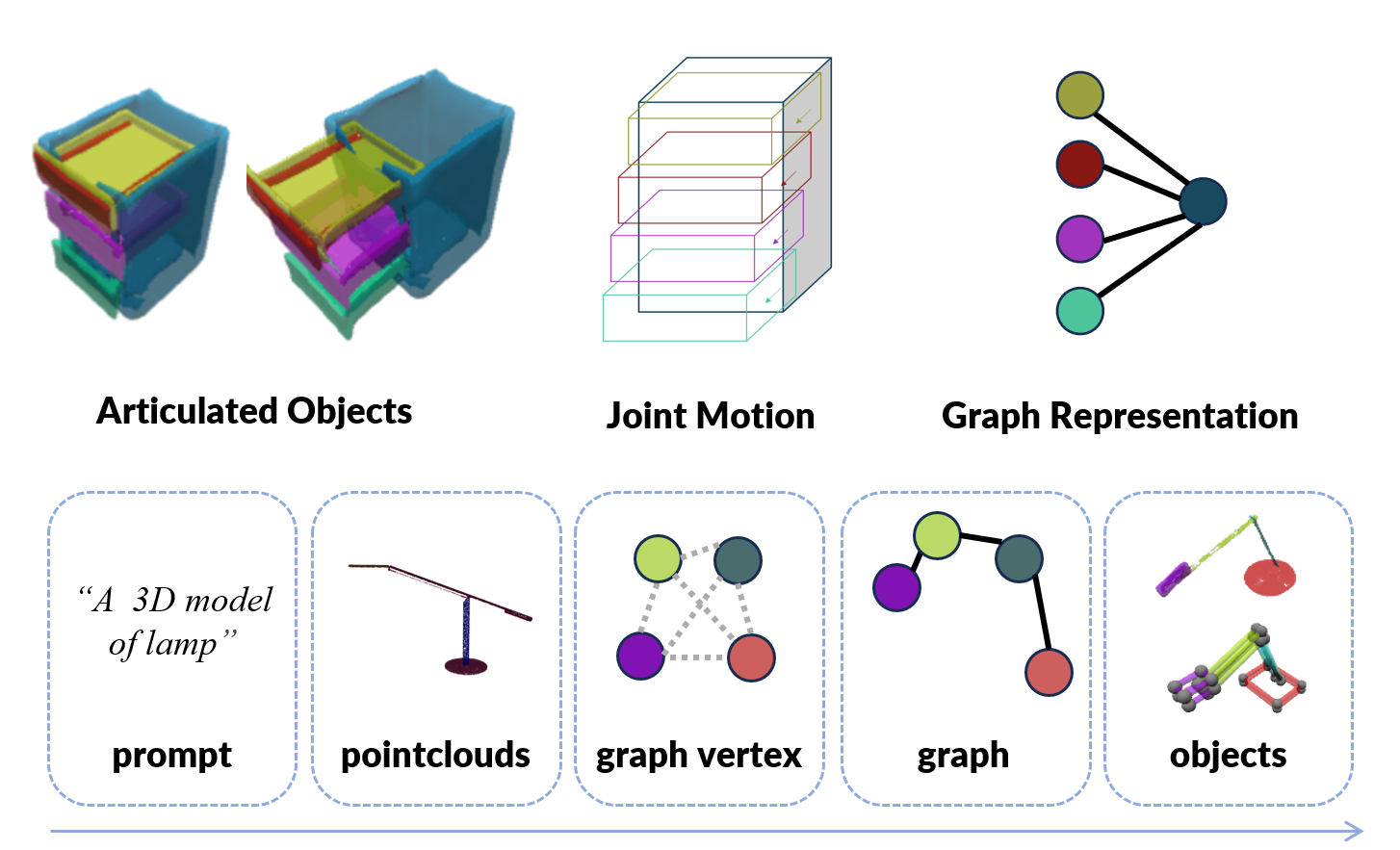}
   \caption{An articulated object consists of several parts and joints. The joints connect different parts and allow them to move. As the joints in most real-world articulated objects are screws with at most one prismatic translation and one revolute rotation, they can be represented by a graph. The lower part of the figure is the overall flow of our proposed approach.}
   \label{fig:overview}
\end{figure}

For example, NAP~\cite{NAP} introduces the first 3D deep generative model capable of synthesizing 3D articulated object models. It generates a comprehensive description of an articulated object, including part geometry, articulation graph, and joint parameters from random noise, operating unconditionally. However, this method lacks the flexibility to generate articulated objects in response to specific prompts. To address this limitation, Cage~\cite{cage} extends the task into a conditional setting, where the object category and a part connectivity graph are given in a user-driven manner as conditions during the reverse process, enabling more targeted and controllable object generation. However, the part connectivity graph input remains challenging to control and lacks intuitiveness, posing difficulties when generating specific objects.

In this paper, we aim to extend existing methods by generating articulated objects conditioned on text prompts, building on recent advances in multimodal learning such as CLIP~\cite{clip} and vision-language models \cite{sun2025llapa,zang2025sage}, which have made it possible to generate target objects from textual descriptions. To address the significant data gap between text descriptions and 3D articulated object components, we propose the Generate Articulated Objects via Text Prompt (GAOT) framework, as illustrated in Figure \ref{fig:overview}. Specifically, we utilize a text-to-3D model to generate initial point clouds from text prompts. To refine these point clouds, we employ a hypergraph-based refinement algorithm, integrating joint features extracted from the dataset into the model. This process transforms the refined point clouds into a graph-based representation of articulated objects using our proposed network. To represent object joints, we combine the vertex graph with Gaussian noise, which is processed by a denoiser in the DDPM framework. The resulting edges derived from the diffusion process define the object joints. Ultimately, the constructed graph is rendered into fully articulated 3D objects.

In summary, the main contributions can be summarized as follows:
\begin{itemize}
    \item We present a three-phase framework that can generate articulated objects from a text prompt.
    \item We design a novel part extraction network to refine the point cloud model. Using a hypergraph-based learning method, connective information of the articulated model saved in a graph-based structure can be extracted.
    \item We design a series of experiments and prove that our method is better than existing methods in terms of articulated model generation, and can generate objects that match the prompt words.
\end{itemize}

\section{Related Work}
\label{sec:rela}
\subsection{Articulated Object Generation}

Articulated objects play a crucial role in creating interactive environments for applications such as VR/AR, gaming, film production, and healthcare \cite{Akb-48,SAPIEN}. Recent research has focused on detection \cite{dect1}, estimation \cite{est1}, and reconstruction \cite{reco1} of articulated objects. For instance, Acdc \cite{Acdc} creates digital twins from RGB images, while URDFormer \cite{dt3, dipo} builds articulated simulation environments from RGBD data.

With the rise of diffusion models and their successful application in various fields~\cite{ma2025tuning,wang2024towards,sun2024eggen}, NAP~\cite{NAP} introduced the first generative model for 3D articulated objects, producing detailed structures including part geometry and articulation graphs. However, NAP lacks conditional generation capabilities in response to particular prompts. Cage~\cite{cage} addresses this limitation by introducing a conditional framework using object categories and part connectivity graphs as inputs, though managing these inputs remains complex and unintuitive. ArtFormer \cite{artformer} introduced a signed-distance-function (SDF) shape prior to facilitate the synthesis of high-quality 3D shapes. 

Recent research has also explored generating or reconstructing articulated objects via other conditions. MeshArt~\cite{meshart} generates a high-level articulation-aware object structure and then synthesizes each part's mesh faces based on this structural information. SINGAPO \cite{singapo} generates articulated objects from a single image with an articulated part attributes generation and part mesh retrieval pipeline based on a diffusion model. DIPO~\cite{dipo} is a novel framework for the controllable generation of articulated 3D objects from a pair of images: one depicting the object in a resting state and the other in an articulated state.

In this paper, we generate articulated objects using a three-phase method.
With the fine-tuned point-E model, we generate the point clouds, which are refined by a hypergraph and transferred to a vertex matrix with an MLP. Then the models are represented as a graph, and the vertices are diffused to obtain the edges. The graph is rendered to articulated objects.

\subsection{Text-to-3D Generation}
Text-to-3D generation methods can be broadly categorized into feedback methods and feedforward methods. 
Feedback methods leverage text-to-image models and optimize 3D parameters through distillation losses. For example, DreamFusion \cite{dreamfusion} proposes Score Distillation Sampling (SDS) as an optimizing technique, while the Score Jacobian Chaining (SJC) \cite{sjc} is mathematically equivalent to SDS.
Although these methods offer better 3D consistency, they are expensive in terms of computational time and memory usage.
The feedforward methods employ complex network structures trained on large-scale 3D datasets such as Objaverse \cite{objaverse}, enabling more accurate and visually appealing 3D object generation.
Some approaches \cite{lgm,t23gs} generate 3D objects from text prompts using 3DGS-based \cite{3dgs} architectures. These methods offer faster generation times but face challenges in zero-shot generation due to limited generalization capabilities.

In this paper, we employ a feedforward method, Point-E \cite{pointe}, within architectures to generate point clouds from a text prompt. With the ability to generate high-quality point clouds, we use it as the first step to obtain a coarse 3D model initially.

\begin{figure*}[htp]
  \centering
   \includegraphics[width=\linewidth]{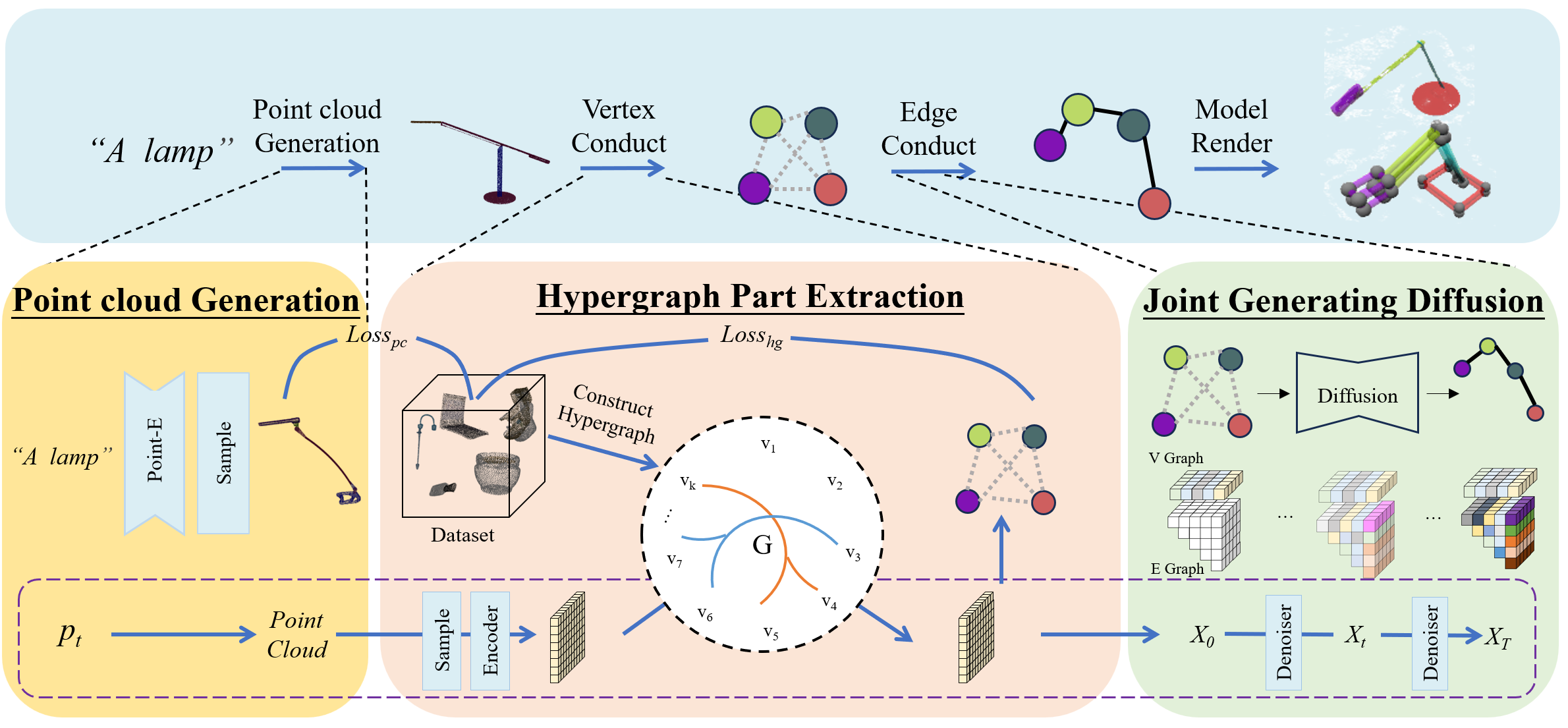}
   \caption{Our framework. Firstly, we generate the point cloud model via a fine-tuned point-E model. The point cloud is refined by a hypergraph constructed from the dataset and then transferred to a vertex matrix with an MLP. Secondly, the models are represented as a graph, in which the parts of objects are transferred to graph vertices and the joints are transferred to graph edges. The graph vertices from the previous stage and random noise edges in matrix form are diffused to obtain complete edges. Finally, the graph is rendered to articulated objects.}
   \label{fig:overall}
\end{figure*}

\section{Method}
\label{sec:method}
Our proposed framework, GAOT, generates articulated objects in a three-phase manner, as illustrated in Figure \ref{fig:overall}. It consists of three main components: point cloud generation, hypergraph part extraction, and joint generating diffusion. Given a text prompt \( p_t \) (\textit{e.g.}, ``a water faucet''), the framework first generates a 3D point cloud representation \( PC_{p_t} \), which is refined into a structured hypergraph representation \( hg(v, e, w)_{p_t} \). Through hypergraph part extraction, the graph vertices \( v_{p_t} \) are obtained, forming the initial articulated object graph \( g_{p_t}(v, e) \). Finally, a diffusion model generates edges \( e_{p_t} \), completing the graph structure. The final articulated object \( g_{p_t}(v_{p_t}, e_{p_t}) \) is rendered, enabling realistic object generation from a text prompt.

\label{subsec:0}

An articulated object can be represented as a graph structure \( G(V, E) \), where \( V = \{v_1, \ldots, v_K\} \) denotes the nodes representing object parts, and \( E = \{e_{ij}\} \) represents the edges corresponding to joints between parts. The number of parts \( K \) defines the nodes, while edges indicate connections based on joints.  Each node \( v_i \) has attributes \( v_i = [o_i, T_{gi}, b_i, f_i] \in \mathbb{R}^{D_v} \), where \( D_v = 1+6+3+F \). Here, \( o_i \in \{0, 1\} \) is a binary indicator of part existence, \( T_{gi} \in \mathbb{R}^6 \) is the SE(3) transformation comprising rotation and translation from local to global coordinates, \( b_i \in \mathbb{R}^3 \) represents the bounding box scale, and \( f_i \) is a latent code encoding part shape from an occupancy autoencoder \cite{autoencoder}. Each edge \( e_{ij} \) has attributes \( e_{ij} = [c_{ij}, p_{ij}, r_{ij}] \in \mathbb{R}^{D_e} \), where \( D_e = 1+6+4 \). The indicator variable \( c_{ij} \in \{-1, 0, 1\} \) denotes edge existence and chirality, where 0 indicates no edge and \( \pm 1 \) specifies edge chirality. The Pl\"ucker coordinate \( p_{ij} \) defines the joint axis globally, while \( r_{ij} \in \mathbb{R}^{2 \times 2} \) specifies joint state ranges for prismatic and revolute motion constraints. This complete graph representation effectively models articulated objects and their structural relationships.

Our goal is to train a generative model $g(G(V,E)|p_t, \theta)$, which can generate articulated objects represented by graph $G(V,E)$ from text prompt $p_t$ (\textit{e.g.}, ``a water faucet''). With text prompt $p_t$, we first generate point clouds $PC_{p_t}$  and refine them with $hg(v,e,w)_{p_t}$. With hypergraph part extraction, we get the articulated object graph vertices $v_{p_t}$ in $g_{p_t}(v,e)$. Finally, we diffuse the $g_{p_t}(v_{p_t},e)$ to get the $e_{p_t}$ and render the $g_{p_t}(v_{p_t},e_{p_t})$ into articulated objects.

\subsection{Point Cloud Generation}
\label{subsec:1}

Given a text prompt \( p_t \), we embed the prompt using CLIP \cite{clip} and generate the corresponding point cloud \( PC_{p_t} \) using a transformer-based network \cite{tranformer}. We adopt the same structure as Point-E \cite{pointe} and fine-tune the model on all parameters using the PartNet-Mobility dataset. The model is optimized using the Mean Squared Error (MSE) loss between the target and predicted point clouds:
\begin{align}
    \text{Loss}_{pc} = \sum_i (\text{MSE}([x,y,z], [x',y',z']) 
    \notag
    \\+\text{MSE}([R,G,B], [R',G',B'])),
\end{align}
where $( x, y ,z ,R ,G ,B ) \in P$ and $(x', y',z',R',G',B') \in P'$, $P$ and $P'$ are point clouds of dimension 10000 $\times$ 6.

In the PartNet-Mobility dataset, the color of point clouds reflects joint information with higher granularity. Model parts are segmented by movable joints, resulting in fewer parts than in PartNet. Points within each part are assigned the same color to represent joint components visually.

Since the PartNet-Mobility dataset only provides categorical labels (\textit{e.g.}, ``refrigerator''), we expand text prompts to more specific descriptions such as ``a refrigerator model X.'' Using unique and non-repetitive prompts helps reduce distribution errors and minimizes model failure cases.

\begin{table*}[tp]
\caption{Main result. We test the methods that generate articulated objects by directly connecting a text-to-3D method or a point cloud segmentation method with NAP.}
  \begin{center}

      \begin{tabular}{l|ccc|ccc}
\hline

           &  \multicolumn{3}{c|}{Simple prompt} & \multicolumn{3}{c}{Complex prompt} \\
        & COV$\uparrow$  & MMD$\downarrow$ & 1-NNA $\downarrow$ & COV$\uparrow$  & MMD$\downarrow$ & 1-NNA$\downarrow$\\
\hline
        
        Point-E\&NAP   & 0.4775$\pm$0.02 & 0.0451$\pm$0.02   & 0.8315$\pm$0.06   & 0.2857$\pm$0.01  & 0.0639$\pm$0.02 & 0.9115$\pm$0.03 \\
        Shap-E\&NAP     & 0.4934$\pm$0.04    & 0.0441$\pm$0.01  & 0.8576$\pm$0.04  & 0.3434$\pm$0.01   & 0.0501$\pm$0.01  & 0.8767$\pm$0.05\\
        PointNet++\&NAP  & 0.5038$\pm$0.07   & 0.0394$\pm$0.01 &  0.6934$\pm$0.02 & 0.3656$\pm$0.01   & 0.0568$\pm$0.01    & 0.8004$\pm$0.07  \\
         \textbf{GAOT (Ours)}    & \textbf{0.5262$\pm$0.07}  & \textbf{0.0383$\pm$0.01}   & \textbf{0.6751$\pm$0.07} & \textbf{0.4898$\pm$0.03} & \textbf{0.0429$\pm$0.01}   & \textbf{0.7512$\pm$0.08}\\
\hline
      \end{tabular}
    \label{tab:mainresult}

  \end{center}
\end{table*}

\subsection{Hypergraph Part Extraction}
\label{subsec:2}
The 3D-generated point cloud has a coarse structure and lacks the connective information required for representing articulated models, which is better captured by a graph-based structure. Traditional graph learning methods are limited to pairwise relationships, while hypergraphs offer a more expressive representation by modeling high-level interactions among multiple entities~\cite{qu2025spatially,wang2025hypergraph,qu2025memory,ijcai2025p201,jing2025multi}. Therefore, we adopt a hypergraph learning approach to refine the pattern matrix. Specifically, we use a Hypergraph Neural Network (HGNN) \cite{HGNN} to convert the point cloud representation into a structured graph representation.  

To construct the hypergraph structure \( hg(v, e, w) \), we process all point clouds in the dataset. For each point cloud, we sample 1024 points using Farthest Point Sampling and encode these points into pattern vectors using a pre-trained PointNet \cite{pointnet} encoder. We then apply the K-means algorithm to cluster these vectors into a matrix with dimensions \( C \times 1024 \), where \( C \) represents the number of hypergraph nodes, encoding both parts and their relationships. With the hypergraph structure we construct from all the point cloud models in the dataset.

In the inference procedure, we follow the hypergraph frequency-domain convolution learning method in HGNN. The convolution layer in HGNN can be represented as:
\begin{equation}
\begin{gathered}
    X^{(0)} = X \in \mathbb{R}^{N \times C}, \\
    X^{{l+1}} = \sigma(D_v^{-\frac12} G W D_e^{-1} H^T D_v^{-\frac12} X^{(l)} \Theta),
\end{gathered}
\end{equation}
where $G\in \mathbb{R}^{N \times E}$ is the incidence matrix of $hg(v,e,w)$, $W$ $=$ $diag$$(w_1$, $w_2$, $\dots$ , $w_n)$ , $\Theta$ denotes the learnable parameters.

Given the point cloud generated by our fine-tuned Point-E, we first extract the pattern vector with the encoder. Then we smooth the vectors with HGNN’s Laplacian. Finally, we obtain the vertex matrix $M_v \in \mathbb{R}^{B \times K \times  D_V}$ of the graph-based articulated objects with an MLP.

To precisely generate the vertex matrix, we design the training loss based on preliminary experiments. As mentioned in Section \ref{subsec:1}, each node has an attribute $v_i \in \mathbb{R}^{D_v}$. The parameters $o_i, T_{gi}$ and $b_i$ describe the basic attributes of the parts and play an important role, especially the $o_i$. The tiny errors of $o_i$ may cause a wrong number of parts, and tiny errors of $T_{gi}$ and $b_i$ may cause the part misplacement problem. Therefore, the loss $L_{matrix}$ is the MSE loss of the matrix.
\begin{equation}
    L_{matrix} = \text{MSE}({M_v},{M_v}')
\end{equation} 
The loss \( L_{\text{bbox}} \) combines bounding box loss and existence loss:  
\begin{equation}
 L_{bbox} = \sum_i \big[\text{MSE}(b_i, b_i') + \text{MSE}(o_i, o_i')\big]
\end{equation}
where ${b_i}, {o_i} \in {M_v}$, ${b_i}', {o_i}' \in {M_v}'$.
The loss \( L_{\text{exist}} \) is defined as the product of $o_i$  and other parameters:
\begin{equation}
    L_{exist} = \text{MSE}(o_i \cdot M_v, o_i' \cdot M_v')
\end{equation}

In summary, the loss of hypergraph part extraction is:
\begin{equation}
    Loss_{hg} = \lambda_1 L_{matrix} + \lambda_2 L_{bbox} + \lambda_3 L_{exist}
\end{equation}
where $\lambda_1$, $\lambda_2$, and $\lambda_3$ are the balance hyperparameters.

The overall loss is the combination of point cloud loss $ L_{pc}$ and hypergraph loss $L_{hg}$. 

\subsection{Joint Generating Diffusion}
\label{subsec:3}
For the Graph to Articulated Objects, we diffuse the vertex matrix $M_v$ we obtained with dimension $B \times K \times D_v$ and a Gaussian noise edge matrix $M_e$ with dimension $B \times K \times K(K-1)/2 \times D_e$ together. 
 
To fully diffuse the edge matrix, we randomly add -1 or 1 to the first column. And during the diffusion process, we refresh the $M_v'$ with $M_v$, for more stable performance of the final articulated objects.

A notable property is that $M_t$ at arbitrary timesteps $t$ can be directly sampled from $M_0$ with $q(M_t|M_0) = N(\sqrt{\bar{\alpha}_t}M_0, (1-\bar{\alpha}_t)I)$,
where $\alpha_t := 1-\beta_t$ and $\bar{\alpha}_t := \prod_{s=1}^t \alpha_s$. 

Starting from a standard Gaussian distribution $x_T\sim N( \mathbf{0}, I)$, we aim to learn a series of Gaussian transitions $p_\theta(x_{t-1} | x_t)$ parameterized by a neural network with learnable weights $\theta$ that gradually removes the noise. 

Following \cite{DDPM}, we set $\Sigma_\theta(x_t,t) = \sigma_t^2I$ and :
\begin{equation}
\begin{gathered}
\label{eq:diffusion:langevin}
    M_{t-1} = \frac{1}{\sqrt{\alpha_t}} \left( M_t - \frac{1-\alpha_t}{\sqrt{1-\bar{\alpha}}_t} \epsilon_\theta(M_t,t) \right) + \sigma_tz, 
\end{gathered}
\end{equation}
where $z \sim  N(\mathbf{0},I)$, and $\epsilon_\theta(M_t,t)$ is a learnable network approximating the per-step noise on $M_t$.

The graph $M'=(M_v, M_e')$ generated from the diffusion process is a fully connected (complete) graph.
This graph is converted to a tree $g_{p_t}(v, e)$ in the final post-processing by computing the Minimum Spanning Tree, which yields a single tree since the graph is complete. The weight of the graph edge is $o_i \in M_e'$ . Finally, we predict joint coordinates and render the articulated objects from the graph.

\begin{figure*}[tp]
  \centering
   \includegraphics[width=\linewidth]{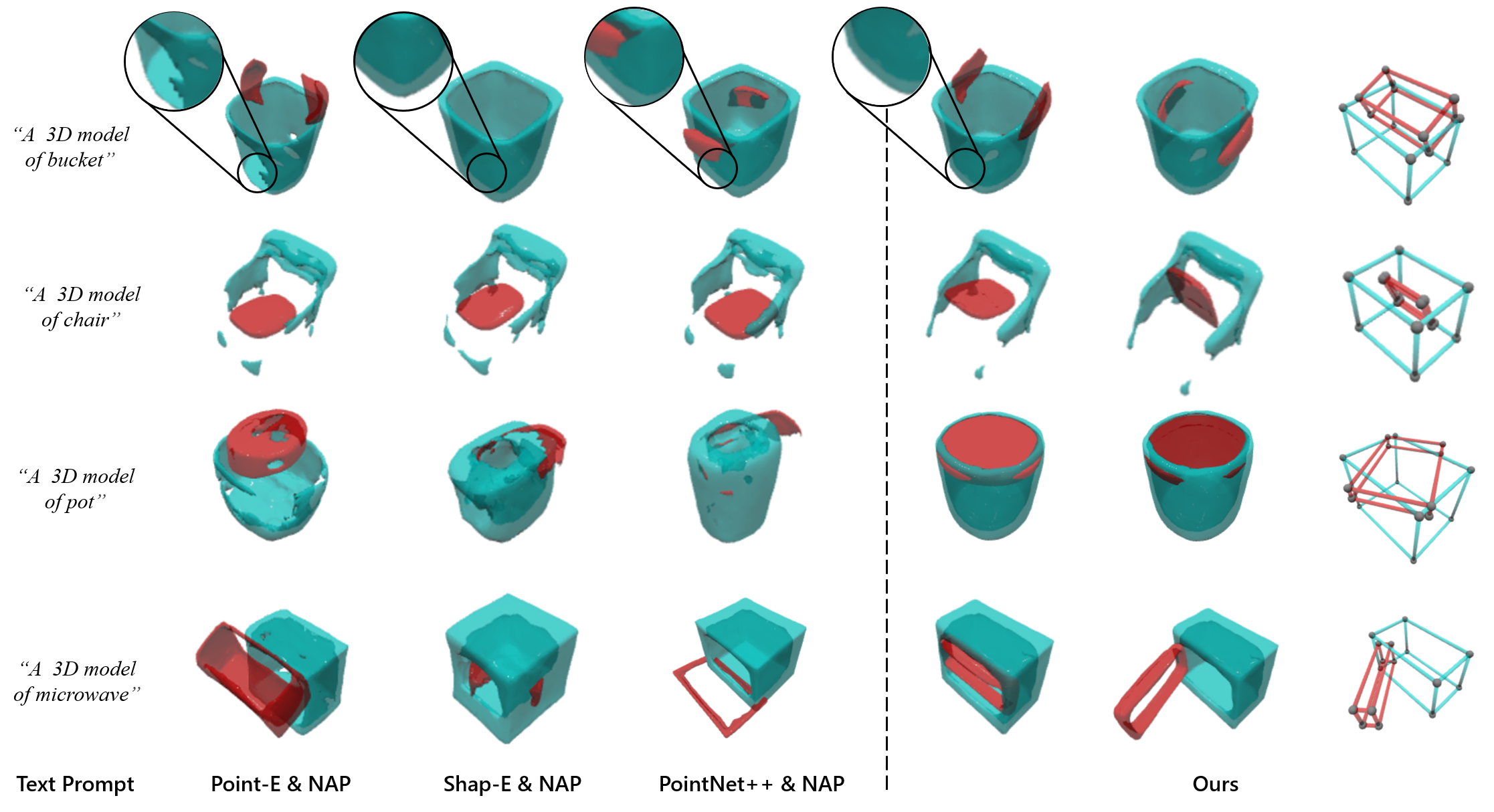}
   \caption{We benchmark our approach against previous methods from the same text prompt. Qualitative comparison shows that our approach can generate objects with complete geometry and rational joint movement. Objects generated using our method are more refined in the details.
}
   \label{fig:compare}
\end{figure*}

\section{Experiments}
\label{sec:exper}
In this section, we elaborate on our experiments conducted to validate the effectiveness of the proposed GAOT approach. Specifically, we benchmark our method against previous methods in the domain of articulated objects generations. A series of comparison experiments and ablation studies are conducted to evaluate the significance of crucial components, including hypergraph learning and diffusion models. 

\subsection{Implementation Details}
We fine-tuned the Point-E model and trained our text-to-graph model on the PartNet-Mobility dataset. For text-to-graph model training, we excluded models with a part count greater than 8. For the data in the dataset that lacks a point cloud model, we used Open3D to extract the point cloud model. All experiments were conducted using an NVIDIA A800 GPU. The basic experimental setup was as follows: 3000 iterations, $K=8$, $F=128$, $C=64$, $\lambda_1=0.4$, $\lambda_2=0.4$, and $\lambda_3=1$. Since $K=8$, we filtered models with more than 8 parts in the dataset.
In the point cloud generation phase, we sampled $PC_{p_t}$ into 1024 points using the Farthest Point Sampling algorithm to accelerate training.
We trained the network with a learning rate of 1e-4 and a learning rate scheduler $step=20$, $\gamma=0.7$ for the Adam optimizer. Training for 3000 iterations takes 24 hours with a batch size of 64.

To test the performance of our model,  we used a set of prompts including both in-dataset and out-of-dataset prompts. The format of a simple prompt is  ``a XXX'', and the complex prompt is ``a 3D XXX model type Y'' where ``Y'' is a random integer to control variations in the generation of articulated objects.

\subsection{Results and Analysis}
We benchmark our approach against three baseline methods, as shown in Table \ref{tab:mainresult}.
Following NAP \cite{NAP}, we use Instantiation Distance (ID) to measure the distance between two articulated objects, considering both part geometry and overall motion structure. Specifically, we randomly sample points and compute the Chamfer distance. We adopt the following three metrics for evaluation: Minimum Matching Distance (MMD) to measure generation quality, Coverage (COV) to test the fraction of the reference set covered, and 1-Nearest Neighbor Accuracy (1-NNA) to measure the distance between distributions via 1-NN classification accuracy.

We test baseline methods that directly connect text-to-3D models with NAP. For text-to-3D generation, we use Point-E and Shap-E, and for point cloud to articulated object conversion, we use PointNet++. These baselines use a simple MLP to obtain $M_v$ and then apply diffusion to $M_e$. 
The results are shown in Table \ref{tab:mainresult}. Quantitative results demonstrate the superiority of our method across all three evaluation metrics, particularly in COV.

We also compare our method with ArtFormer \cite{artformer}, which generates diverse 3D articulated objects conditioned on text or images. Since ArtFormer uses 6 categories (Storage Furniture, Safe, Oven, USB, Bottle, and Washer) from PartNet-Mobility, we follow the same setup. Results in Table \ref{tab:comresult} show that our method achieves comparable performance.

\begin{table}[h]
    \caption{Comparison results with ArtFormer on a subset of PartNet-Mobility. The best results are highlighted in \textbf{bold}.}
  \begin{center}
      \begin{tabular}{lccc}
        \hline
        & COV$\uparrow$  & MMD$\downarrow$ & 1-NNA $\downarrow$\\
        \hline
        ArtFormer-NAPSP & 0.4831   & 0.0375 & 0.8315   \\
        ArtFormer &   0.5213 & \textbf{0.0292} & \textbf{0.5266}   \\
          \textbf{GAOT(Ours)}  & \textbf{0.5262$\pm$0.07} & 0.0319$\pm$0.01   & 0.6012$\pm$0.06 \\
        \hline
      \end{tabular}
    \label{tab:comresult}
  \end{center}
\end{table}

\begin{figure}[tp]
  \centering
   \includegraphics[width=\linewidth]{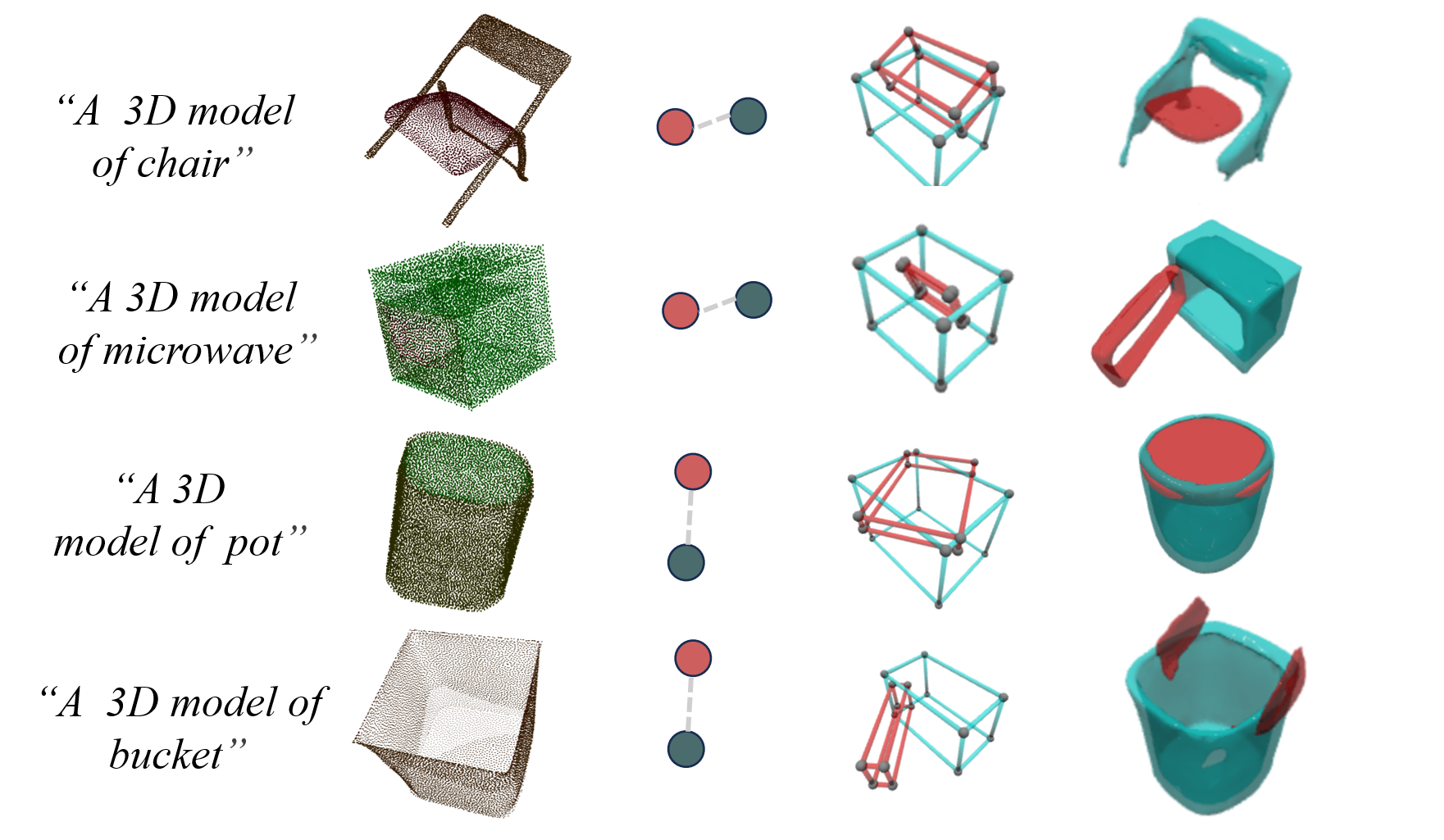}
    \caption{The articulated object generation process. First, point clouds with colored structural information are generated. Then, through hypergraph part extraction, we obtain parts represented as graph vertices. Using a diffusion model, we generate joints and render the articulated objects.}
   \label{fig:visresult}
\end{figure}

From visual results (Figure \ref{fig:compare}), we see that our method outperforms all baselines. The Point-E \& NAP method tends to generate objects incompletely, and the other two methods tend to generate objects with incorrect shapes or joints. 
We visualize our result and generation process in Figure \ref{fig:visresult}. The point clouds with colored structure information are generated first and are well-shaped, allowing different parts to be distinguished by color. With the graph vertices and formed edges, we generate the articulated objects. Based on our results, we can find that the articulated objects we generate have a complete geometry, while the position and the rotation direction of the joints match the situation in the real world. 

\begin{figure}[tp]
  \centering
   \includegraphics[width=0.95\linewidth]{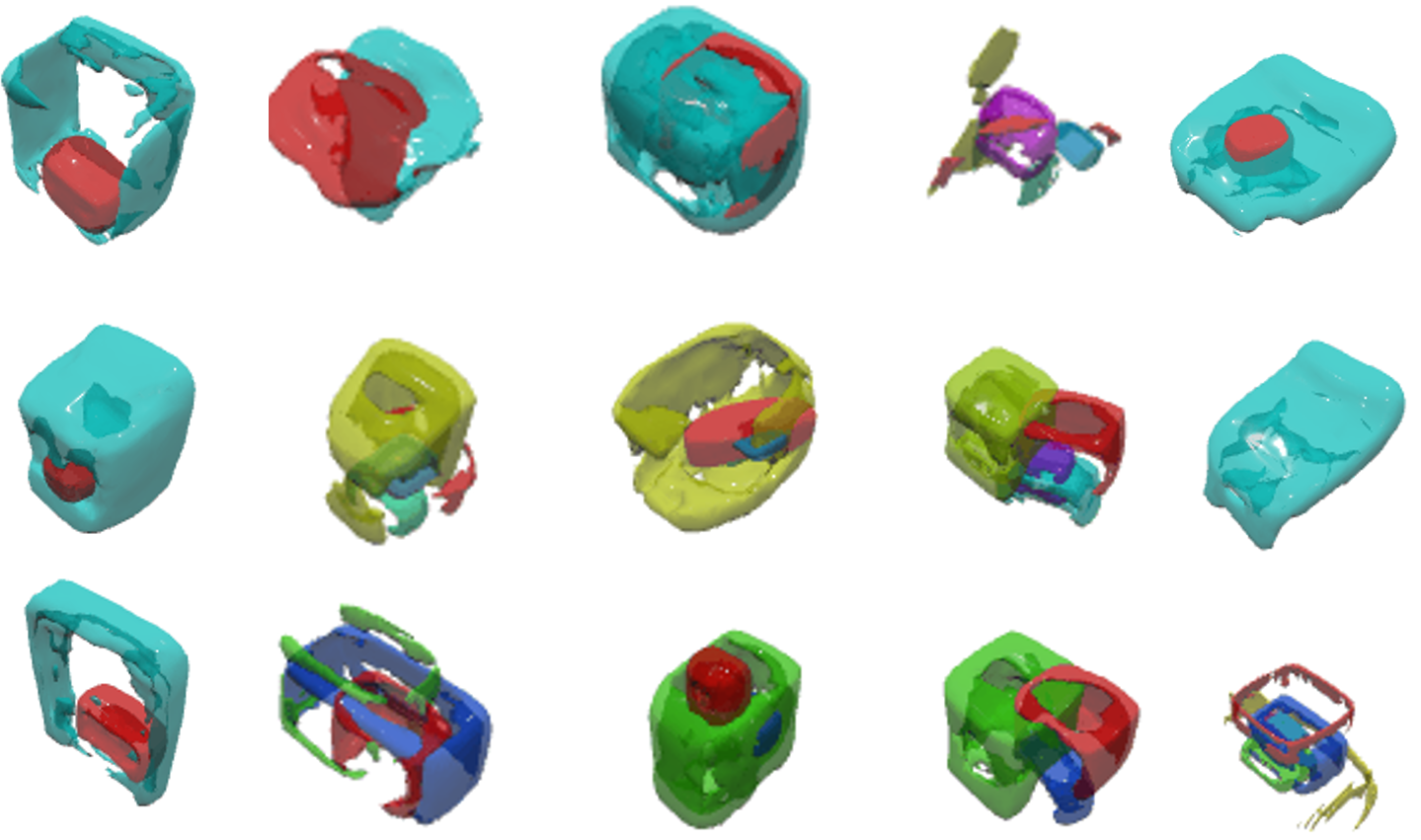}
   \caption{Several failure cases in terms of both generation quality and controllability.}
   \label{fig:fail}
\end{figure}

Some failure cases are shown in Figure \ref{fig:fail}. Articulated objects may exhibit issues such as: 1) misaligned parts (\textit{e.g.}, upper and lower cabinet doors), 2) incomplete shapes (\textit{e.g.}, broken parts), and 3) incorrect joint movements (\textit{e.g.}, unrealistic door or handle motions). The main cause is minor deviations in key parameters of the graph structure.

\begin{table}[t]
    \caption{Ablation result of hypergraph learning.}

  \begin{center}
      \begin{tabular}{lccc}
\hline
        & COV$\uparrow$  & MMD$\downarrow$ & 1-NNA $\downarrow$\\
\hline
        w/o hypergraph & 0.5238$\pm$0.07   & 0.0394$\pm$0.01 & 0.7015$\pm$0.03   \\
        w graph (GCN) & 0.5644$\pm$0.05   & 0.0362$\pm$0.01 & 0.6241$\pm$0.04  \\
         \textbf{HGNN}    & \textbf{0.6562$\pm$0.07} & \textbf{0.0319$\pm$0.01}   & \textbf{0.6012$\pm$0.06} \\
\hline
      \end{tabular}
    \label{tab:ablation1}
  \end{center}
\end{table}

\begin{table}[t]
    \caption{Ablation result of diffusion model.}

  \begin{center}
      \begin{tabular}{lccc}
        \hline
        & COV$\uparrow$  & MMD$\downarrow$ & 1-NNA $\downarrow$\\
        \hline
        $L_{matrix}$ & 0.5822$\pm$0.06   & 0.0334$\pm$0.01 & 0.7263$\pm$0.05   \\
        $L_{matrix}$+ $L_{bbox}$ & 0.5012$\pm$0.08   & 0.0504$\pm$0.02 & 0.8761$\pm$0.02   \\
        $L_{matrix}$+ $L_{exist}$ & 0.6004$\pm$0.03   & 0.0343$\pm$0.01 & 0.6401$\pm$0.03   \\
        $L_{bbox}$+ $L_{exist}$ & 0.6445$\pm$0.05   & 0.0349$\pm$0.01 & 0.6315$\pm$0.04  \\
          \textbf{OURS}  & \textbf{0.6562$\pm$0.07} & \textbf{0.0319$\pm$0.01}   & \textbf{0.6012$\pm$0.06} \\
        \hline
      \end{tabular}
    \label{tab:ablation3}
  \end{center}
\end{table}

\subsection{Ablation Study}

To demonstrate the superiority of hypergraph learning, we compare our method against variants without hypergraph learning and with graph learning (GCN). Results in Table \ref{tab:ablation1} show that the hypergraph structure represents better complex joint information in articulated objects and outperforms graph learning.

To demonstrate the effectiveness of our designed loss function, we ablate the loss components in the hypergraph part extraction module. Results in Table \ref{tab:ablation3} show that our loss function leads to more complete and accurate geometric structures with proper bounding boxes, achieving higher MMD performance.

\section{Conclusions}
\label{sec:conclu}
We introduce GAOT, a novel method for generating articulated objects from text prompts. Our approach employs a three-phase generation process that bridges the gap between text descriptions and articulated objects through hypergraph learning, graph representation, and diffusion models. We believe our method can inspire new research directions, and we will focus on generating more detailed and versatile articulated objects in future work.


\bibliographystyle{ACM-Reference-Format}
\bibliography{sample-base}

\end{document}